\documentclass{article} 
\usepackage{iclr2027_conference,times}

\usepackage{amsmath,amsfonts,bm}

\def\eqref#1{equation~\ref{#1}}

\def\1{\bm{1}}

\DeclareMathAlphabet{\mathsfit}{\encodingdefault}{\sfdefault}{m}{sl}
\SetMathAlphabet{\mathsfit}{bold}{\encodingdefault}{\sfdefault}{bx}{n}

\usepackage{hyperref}
\usepackage{url}
\usepackage{graphicx}
\usepackage{amsmath}
\usepackage{amssymb}
\usepackage{booktabs}
\usepackage{caption}
\usepackage{placeins}
\usepackage{booktabs}
\usepackage{caption}
\usepackage{subcaption}
\usepackage{placeins}
\usepackage{float}

\newcommand{\tworowhead}[1]{\smash{\raisebox{-1.35ex}{\strut #1}}}

\title{GeoGSplat: Geometry-Grounded Non-Pixel-Aligned Gaussian Splatting}

\author{%
Jiale Wang$^{1,3*}$ \quad Chen Zhao$^{2*}$ \quad Haoyang Zhang$^3$ \quad Yanhao Wu$^{1,3}$ \\[3pt]
Fei He$^3$ \quad Yanhu Shan$^3$ \quad Rui Wu$^3$ \quad Wei Ke$^1$ \quad Tong Zhang$^{4\ddagger}$ \\[7pt]
{\normalsize
$^1$Xi'an Jiaotong University \quad $^2$EPFL \\[2pt]
$^3$Horizon Robotics \quad $^4$University of the Chinese Academy of Sciences}
}

\iclrfinalcopy 

\renewcommand{\headrulewidth}{0pt}

\makeatletter
\renewcommand{\@maketitle}{%
\vbox{\hsize\textwidth
  \centering
  {\normalfont\LARGE\bfseries \@title\par}
  \vskip 12pt
  {\normalfont\large \@author\par}
  \vskip 20pt
}%
}
\makeatother

\begin{document}

\maketitle

\begingroup
\renewcommand{\thefootnote}{}
\footnotetext[0]{%
\fontsize{8}{10}\selectfont
$^*$Equal contribution.\enspace
$^\ddagger$Corresponding author.}
\endgroup

\begin{figure}[h]
\vspace{-6mm}
\begin{center}
\includegraphics[width=\linewidth]{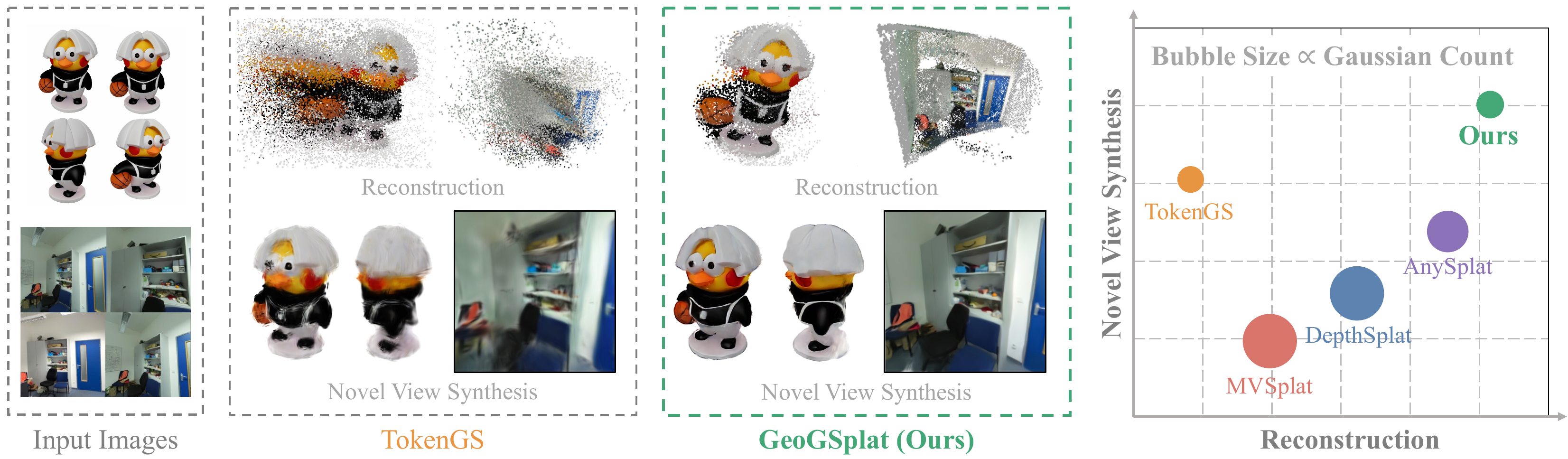}
\end{center}
\caption{\textbf{GeoGSplat} introduces a geometry-grounded non-pixel-aligned Gaussian Splatting paradigm. It achieves high-quality 3D reconstruction and novel view synthesis by incorporating reliable geometric priors through complementary Anchor and Triplane representations. Larger bubble size indicates a larger number of Gaussians.}
\end{figure}

\begin{abstract}
Feed-forward Gaussian Splatting enables efficient novel view synthesis, but pixel-aligned prediction makes the primitive count grow with image resolution and view count, leading to substantial redundancy and increased computational cost. Non-pixel-aligned methods address this limitation by predicting a fixed number of Gaussians. Without explicit ray and depth constraints, these methods yield misplaced Gaussian centers, causing rendering artifacts under large viewpoint changes. To address this issue, we introduce \textbf{GeoGSplat}, a geometry-grounded feed-forward model that explicitly incorporates geometric priors into the non-pixel-aligned framework. A dual-branch encoder constructs \emph{Anchors} and \emph{Triplane} to capture coarse 3D structure and fine-grained scene features, respectively. Subsequently, a hierarchical decoder progressively decodes dense 3D points and Gaussian primitives, enabling 3D reconstruction and novel view synthesis. GeoGSplat surpasses both state-of-the-art pixel-aligned and non-pixel-aligned Gaussian Splatting methods on DTU and NRGBD in terms of novel view synthesis and reconstruction. The advantage of our geometry-grounded paradigm further increases as target viewpoints move farther from the input cameras, demonstrating stronger generalization to large viewpoint changes. \noindent\textbf{Project Page:} \url{https://comedy0913.github.io/projects/geogsplat}
\end{abstract}

\section{Introduction}

\begin{figure}[h]
\begin{center}
\includegraphics[width=\linewidth]{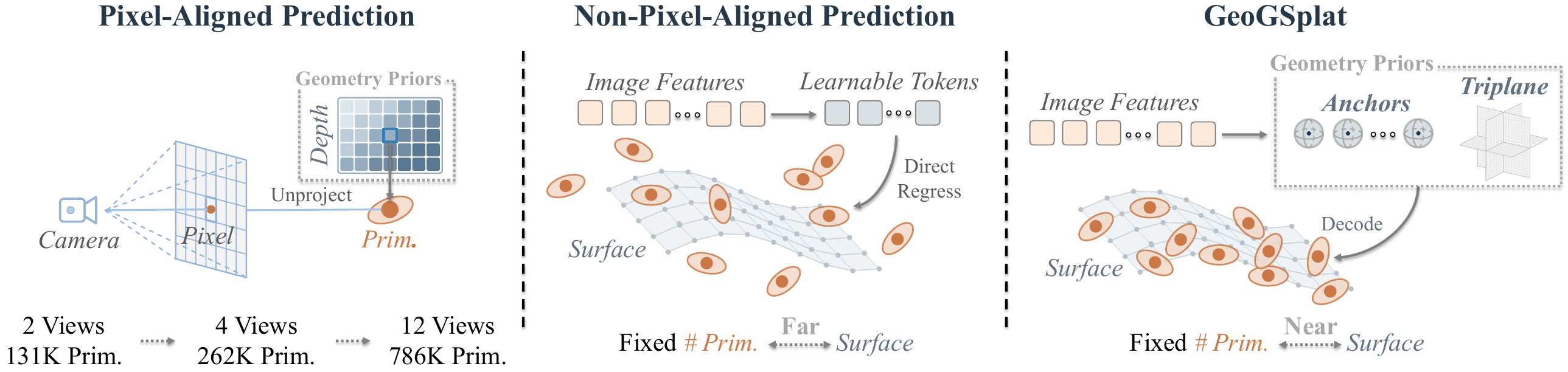}
\end{center}
\caption{\textbf{Comparison of pixel-aligned, non-pixel-aligned, and our GeoGSplat paradigms}. Pixel-aligned methods rely on explicit geometry priors but produce an increasing number of primitives (Prim.) with more input views. Non-pixel-aligned methods predict a fixed number of primitives but lack geometric priors. GeoGSplat combines fixed-size prediction with explicit geometric priors.}
\label{fig:Comparison}
\vspace{-4mm}
\end{figure}

Feed-forward Gaussian Splatting~\citep{charatan2024pixelsplat,chen2024mvsplat,zhang2024gs,tokengs2026} aims to directly predict Gaussian primitives from a set of images, supporting novel view synthesis (NVS) and avoiding expensive per-scene optimization~\citep{kerbl20233d}. Most existing methods~\citep{charatan2024pixelsplat,chen2024mvsplat} conduct pixel-aligned predictions, where each input pixel is associated with a predicted Gaussian primitive. Such methods inherit explicit geometric constraints from per-pixel depth prediction, which restricts Gaussian centers to a geometrically meaningful 3D surface and thus reduces the ambiguity of appearance learning during rasterization. However, this paradigm couples Gaussian primitives to both image resolution and the number of input views, resulting in considerable computational overhead. As illustrated in Fig.~\ref{fig:Comparison}, the number of predicted primitives grows rapidly with the number of input views. Moreover, overlapping regions across images produce multiple Gaussians for the same scene content, introducing substantial redundancy. Some recent methods~\citep{tokengs2026,li2026locusgs,li2026querysplat}, adopt a non-pixel-aligned paradigm that predicts a fixed number of Gaussians. These methods typically decode Gaussians from a set of learnable tokens that aggregate multi-view image features through attention. While this formulation removes the scaling dependency of pixel-aligned prediction, it also discards the geometric constraints provided by per-pixel depth. Primitives are learned from photometric supervision, leaving the Gaussian centers underconstrained and resulting in inaccurate geometry and overfitting to the observed viewpoints. Such Gaussians may produce plausible renderings near the input cameras, but they tend to yield strong rendering artifacts under larger viewpoint changes.

To address this limitation, we introduce \emph{GeoGSplat}, a geometry-grounded non-pixel-aligned framework consisting of a \emph{dual-branch structured encoder} and a \emph{geometry-guided hierarchical decoder}. The encoder captures complementary scene properties through \emph{Anchors} and \emph{Triplane}. Anchors comprise sparse 3D spheres with learnable centers and adaptive radii, providing coarse geometric constraints on the scene structure. Triplane comprises three orthogonal 2D feature maps that encode fine-grained high-dimensional scene features queryable at arbitrary 3D coordinates. Building on the two complementary representations, the decoder progressively converts sparse geometric priors into dense 3D points and Gaussian primitives. We first perform \emph{Anchor-to-Point} decoding by sampling Triplane features within each Anchor region to predict a set of dense 3D points. We then apply \emph{Point-to-Gaussian} decoding, re-querying the Triplane at the decoded 3D points to predict corresponding Gaussian primitives. This hierarchical design establishes interaction between Anchors and Triplane, using Anchor geometry to constrain point reconstruction and employing fine-grained Triplane features to enable geometric refinement and appearance learning. 

We conduct experiments on DTU~\citep{jensen2014large} and NRGBD~\citep{azinovic2022neural} for object-level and scene-level evaluations, respectively. Our GeoGSplat outperforms both non-pixel-aligned and pixel-aligned competitors on these datasets in terms of NVS and 3D reconstruction.
Furthermore, we evaluate GeoGSplat under a challenging setting by progressively moving the target cameras farther from the input cameras. Its performance advantage becomes increasingly pronounced under larger viewpoint changes, showing stronger robustness and generalization under large viewpoint changes enabled by our geometric priors.
Our contributions are summarized below.
\begin{itemize}
    \item We introduce a \textbf{geometry-grounded non-pixel-aligned paradigm} for feed-forward Gaussian Splatting that incorporates explicit geometric priors, enabling high-quality NVS and 3D reconstruction in a fixed Gaussian budget.

    \item We propose a \textbf{dual-branch structured encoder} that constructs Anchors and Triplane to capture complementary coarse 3D structure and fine-grained high-dimensional scene features, respectively.

    \item We develop a \textbf{geometry-guided hierarchical decoder} that progressively decodes dense 3D points and Gaussian primitives by querying Triplane features under Anchor-based geometric constraints.
\end{itemize}

\section{Related Works}
\textbf{Pixel-aligned feed-forward Gaussian Splatting.}
Significant progress has been made in feed-forward Gaussian Splatting~\citep{charatan2024pixelsplat,chen2024mvsplat,zhang2024gs,tokengs2026,szymanowicz2024splatter}, predicting Gaussian primitives from images without per-scene optimization. Most existing approaches follow a pixel-aligned mechanism~\citep{chen2024mvsplat,xu2025depthsplat,charatan2024pixelsplat}, predicting a Gaussian primitive for each image pixel. Such predictions benefit from explicit geometric constraints imposed by the corresponding camera rays and per-pixel depths, which restrict Gaussian centers to plausible 3D locations. However, the pixel-aligned pipeline ties the number of Gaussians to the input image resolution and the number of views, leading to considerable redundancy and computational cost. 

\textbf{Non-pixel-aligned models.} 
Non-pixel-aligned methods remove correspondences between image pixels and Gaussian primitives, instead predicting Gaussians from aggregated image features or scene representations~\citep{wang2026volsplat,zhang2026anchorsplat,li2026locusgs,tokengs2026,zhao2026rayzer,moreau2026off,li2026querysplat}. However, removing pixel-to-Gaussian correspondences also eliminates the geometric constraints provided by per-pixel ray and depth. In this context, Gaussian centers tend to be inaccurately placed, leading to suboptimal solutions that overfit the observed viewpoints. While these primitives may still produce plausible renderings for novel views close to the input cameras, their geometric errors become increasingly exposed under large viewpoint changes, resulting in strong rendering artifacts. To tackle this issue, we introduce a new non-pixel-aligned paradigm that injects explicit geometric priors, enabling more robust NVS.

\textbf{Scene Representations.}
Existing feed-forward methods adopt various scene representations~\citep{yao2018mvsnet,gu2020cascade,yu2021pixelnerf,wang2021ibrnet,chen2021mvsnerf,jiang2025rayzer,johari2022geonerf,wang2024dust3r,leroy2024grounding,wang2025vggt,xiang2026native}. These representations play a critical role in Gaussian prediction. For instance, voxels and anchors constrain the spatial locations at which primitives are predicted~\citep{lu2024scaffold,fridovich2022plenoxels}; Triplane~\citep{hong2024lrm,zhang2024gs,chan2022efficient} provides structured representations, where fine-grained high-dimensional scene features can be encoded and queried. Motivated by the complementarity of these two representations, GeoGSplat combines Anchors and Triplane, using Anchors to provide coarse geometric constraints and Triplane to supply fine-grained scene features.

\section{Method}

\subsection{Preliminaries}
Given $N_I$ posed images $\{(\mathbf{I}_i,\boldsymbol{\pi}_i)\}_{i=1}^{N_I}$, feed-forward Gaussian Splatting predicts a set of 3D Gaussian primitives in a single pass. Here, $\mathbf{I}_i\in\mathbb{R}^{H\times W\times3}$ is an RGB image and $\boldsymbol{\pi}_i$ denotes its camera parameters. The predicted Gaussian set $\mathcal{G}$ is defined as:
\begin{equation}
\mathcal{G}=\{\mathbf{g}_j\}_{j=1}^{N}, \qquad
\mathbf{g}_j=(\boldsymbol{\mu}_j,\alpha_j,\mathbf{s}_j,\mathbf{q}_j,\mathbf{c}_j),
\label{eq:gaussian_set}
\end{equation}
where $N$ denotes the number of predicted Gaussians, $\boldsymbol{\mu}_j$, $\alpha_j$, $\mathbf{s}_j$, $\mathbf{q}_j$, and $\mathbf{c}_j$ represent the center, opacity, scale, rotation, and spherical-harmonic coefficients of $\mathbf{g}_j$~\citep{kerbl20233d}.

\textbf{Pixel-aligned prediction.} For each pixel $\mathbf{u}=(x,y)$ in image $\mathbf{I}_i$, pixel-aligned methods~\citep{chen2024mvsplat} predict a depth $z$ and obtain its Gaussian center $\boldsymbol{\mu}$ by camera unprojection
$\boldsymbol{\mu}=\operatorname{Unproj}(\mathbf{u},z,\boldsymbol{\pi}_i)$, where $\operatorname{Unproj}(\cdot)$ indicates the 2D to 3D unprojection. Since each pixel is mapped to one Gaussian, the total number of Gaussians is $N=N_I\times H\times W$ and therefore grows with both the input resolution and the number of views.

\textbf{Non-pixel-aligned prediction.} Non-pixel-aligned methods~\citep{tokengs2026,li2026querysplat} aggregate the input images into multi-view features $\mathbf{F}_{\mathrm{img}}$ and use $N_Q$ learnable tokens $\mathbf{Q}$ to directly decode Gaussians as $\mathcal{G}=f(\mathbf{F}_{\mathrm{img}},\mathbf{Q})$, where $N=N_Q \times N_{\mathcal{G}/Q}$ and $N_{\mathcal{G}/Q}$ denotes the number of Gaussians predicted per token. This fixed-count formulation enables efficient prediction, but discards the geometric priors that have proven effective in pixel-aligned models.

\begin{figure}[t]
    \centering
    \includegraphics[width=\linewidth]{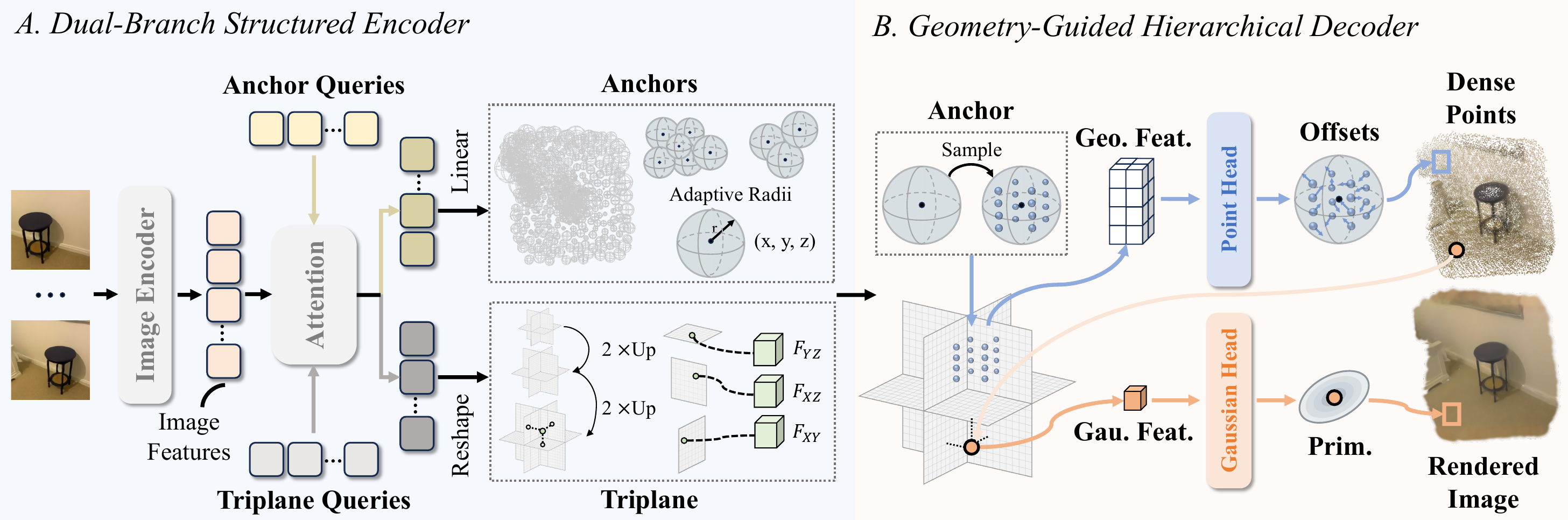}
    \caption{\textbf{Pipeline of GeoGSplat.} Image features are encoded in parallel into \emph{Anchors} and \emph{Triplane}. Anchors comprise sparse 3D spheres with learnable centers and adaptive radii, accounting for the coarse scene structure; Triplane consists of three orthogonal 2D feature maps that encode fine-grained high-dimensional scene features. \emph{Anchor-to-Point Decoding} (blue) gathers Triplane features within each Anchor region to predict offsets and recover dense 3D points; \emph{Point-to-Gaussian Decoding} (orange) re-queries the Triplane at these points to predict the corresponding Gaussian primitives.}
    \label{fig:overview}
    \vspace{-3mm}
\end{figure}

To retain a fixed Gaussian count while keeping explicit geometric priors, we introduce GeoGSplat, a geometry-grounded non-pixel-aligned prediction paradigm. As shown in Fig.~\ref{fig:overview}, GeoGSplat first constructs
$M$ Anchors $\mathcal{A}$ and a Triplane $\mathcal{T}$ from $\mathbf{F}_{\mathrm{img}}$ through a dual-branch structured encoder. Subsequently, a geometry-guided hierarchical decoder is proposed to expand each Anchor into $K$ dense 3D points and decode the corresponding $K$ Gaussians. As $M$ and $K$ are fixed, $N=M \times K$ remains independent of $(N_I,H,W)$.

\subsection{Dual-Branch Structured Encoder}
\label{sec:representations}
\textbf{Image Feature Extraction.} For each view, RGB and their corresponding Pl\"ucker ray~\citep{plucker1865new} embeddings are summed and processed by an image encoder, implemented as a standard Vision Transformer~\citep{dosovitskiy2020vit}:
\begin{equation}
\mathbf{F}_{\mathrm{img}}
=
\operatorname{Encoder}\!\left(
\left[
\operatorname{PE}_{\mathrm{rgb}}(\mathbf{I}_i)
+\operatorname{PE}_{\mathrm{ray}}(\operatorname{Ray}(\boldsymbol{\pi}_i))
\right]_{i=1}^{N_I}
\right),
\quad
\mathbf{F}_{\mathrm{img}}\in\mathbb{R}^{N_In_p\times C}.
\label{eq:image_encoding}
\end{equation}
where $n_p$ is the number of patches per view, $\operatorname{PE}_{\mathrm{rgb}}(\cdot)$ and $\operatorname{PE}_{\mathrm{ray}}(\cdot)$ indicate standard ViT patch embedding modules applied to RGB patches and Pl\"ucker rays, respectively. 

\textbf{Dual-Branch Feature Aggregation.} To extract features for Anchor and Triplane construction, we introduce two sets of learnable queries, denoted as $\mathbf{Q}^{A}\in\mathbb{R}^{M\times C}$ and $\mathbf{Q}^{T}\in\mathbb{R}^{n_T\times C}$, respectively, where $n_T=3S_0^2$ and $S_0$ is the initial plane resolution. For the Triplane branch, we additionally introduce spatial positional embeddings $\mathbf{P}^{T}\in\mathbb{R}^{n_T\times C}$ to encode the 2D grid coordinates and identities of the three orthogonal planes. The two sets of queries aggregate the image features through a shared attention module $\operatorname{Att}(\cdot)$~\citep{carion2020end} as
\begin{equation}
\begin{aligned}
\mathbf{F}^{A}&=\operatorname{Att}(\mathbf{Q}^{A},\mathbf{F}_{\mathrm{img}}),
&\mathbf{F}^{A}&\in\mathbb{R}^{M\times C},
&\mathbf{F}^{T}&=\operatorname{Att}(\mathbf{Q}^{T}+\mathbf{P}^{T},\mathbf{F}_{\mathrm{img}}),
&\mathbf{F}^{T}&\in\mathbb{R}^{n_T\times C}.
\end{aligned}
\label{eq:feature_decoder}
\end{equation}
The resulting $\mathbf{F}^{A}$ and $\mathbf{F}^{T}$ are subsequently used to construct Anchors and Triplane.

\textbf{Anchor Construction.} To provide coarse geometric constraints on the scene structure, we construct $M$ Anchors $\mathcal{A}=\{(\mathbf{p}_i,r_i)\}_{i=1}^{M}$ from $\mathbf{F}^{A}$. Each Anchor is represented by a 3D sphere with center $\mathbf{p}_i$ and adaptive radius $r_i$. Specifically, we initialize the Anchor centers from a set of reference points $\mathcal{Z}=\{\mathbf{z}_i\}_{i=1}^{M}$ with broad spatial coverage, then refine them according to $\mathbf{F}^{A}$. The reference points $\mathcal{Z}$ are randomly sampled once from the normalized 3D volume and kept fixed throughout training. We predict an offset from $\mathbf{F}^{A}_i$ with a linear projection $Linear(\cdot)$. The offset is then applied to the corresponding reference point $\mathbf{z}_i$ to obtain the center $\mathbf{p}_i$:
\begin{equation}
\mathbf{p}_i=\tanh\!\left(\mathbf{z}_i+\text{Linear}(\mathbf{F}^{A}_i)\right),
\qquad \mathbf{p}_i\in(-1,1)^3,
\label{eq:anchors}
\end{equation}
where $\tanh(\cdot)$ keeps the predicted centers within the normalized scene volume. Given the refined Anchor centers, we further determine their spatial extents using adaptive radii to accommodate the non-uniform spatial distribution of 3D scene content. We initialize the radius $r_i$ using the median distance from $\mathbf{p}_i$ to the four nearest Anchor centers. The radius is then adjusted with a bounded factor predicted from $\mathbf{F}^{A}_i$ as:
\begin{equation}
r_i=d_i(\gamma_{\min}+(\gamma_{\max}-\gamma_{\min})\,\sigma(\text{Linear}(\mathbf{F}^{A}_i))),
\qquad r_i\in\mathbb{R}_{+},
\label{eq:radius}
\end{equation}
where $d_i$ defines the median distance, $(\gamma_{\min},\gamma_{\max})$ bound the learned multiplier, and $\sigma(\cdot)$ is the sigmoid function. $d_i$ adapts the Anchor extent to the spatial distribution of scene geometry, while the learned factor makes it responsive to image content. Together, these Anchors form a coarse 3D representation of the scene observed in the input images.

\textbf{Triplane Construction.} To complement the coarse structure with fine-grained features, we build Triplane from $\mathbf{F}^{T}$ through progressive refinement. First, we reshape $\mathbf{F}^{T}\in\mathbb{R}^{3S_0^2\times C}$ to an initial Triplane $\mathbf{T}_{S_0}\in\mathbb{R}^{3\times S_0\times S_0\times C}$. Two refinement stages are then applied to increase the plane resolution from $S_0$ to $S = 4S_0$. Each refinement comprises bilinear upsampling by a factor of two $\operatorname{Up}_{2}(\cdot)$ and a Triplane refinement block $\mathcal{R}_s(\cdot,\cdot)$:
\begin{equation}
\mathbf{T}_{2s}=\mathcal{R}_s\!\left(\operatorname{Up}_{2}(\mathbf{T}_s),\mathbf{F}_{\mathrm{img}}\right),
\qquad \mathbf{T}_{2s}\in\mathbb{R}^{3\times 2s\times 2s\times C^{'}},
\label{eq:triplane_refine}
\end{equation}
where $s$ and $C^{'}$ denote the plane resolution and feature dimension, respectively. $\mathcal{R}_s(\cdot,\cdot)$ combines residual convolution~\citep{he2016deep}, cross-plane feature fusion~\citep{wang2023rodin}, and image-conditioned cross-attention. The final Triplane is defined as $\mathcal{T}=\{\mathbf{T}^{xy},\mathbf{T}^{xz},\mathbf{T}^{yz}\},\mathbf{T}\in\mathbb{R}^{S\times S\times C_T}$. Given 3D coordinates $\mathbf{x}\in[-1,1]^3$, we project them onto each plane and sample features at the corresponding 2D locations. The Triplane features $\phi(\mathbf{x})$ are then obtained via concatenation. Such a query process is defined as
\begin{equation}
\phi_a(\mathbf{x})
=
\operatorname{Bilinear}(\mathbf{T}^{a},\operatorname{Proj}(\mathbf{x})),
\quad
\phi(\mathbf{x})
=
[\phi_{xy}(\mathbf{x});\phi_{xz}(\mathbf{x});\phi_{yz}(\mathbf{x})]
\in\mathbb{R}^{3C_T}.
\label{eq:sample}
\end{equation}
where $a\in\{xy,xz,yz\}$ identifies a plane, $\operatorname{Proj}(\cdot)$ maps the 3D coordinates to 2D coordinates on a plane, and $\operatorname{Bilinear}(\cdot,\cdot)$ samples the corresponding features by bilinear interpolation. The queried features are utilized in the following decoder for reconstruction and Gaussian primitive prediction.

\subsection{Geometry-Guided Hierarchical Decoder}
\label{sec:queries}

To enforce geometric priors throughout decoding, we propose a hierarchical decoder with two stages: Anchor-to-Point decoding reconstructs dense 3D geometry, and Point-to-Gaussian decoding builds Gaussian primitives upon the recovered geometry for appearance modeling.

\textbf{Anchor-to-Point Decoding.} To decode dense points within each Anchor, we initialize reference points and then refine them using scene features from the Triplane. Similar to Anchor construction, these points are randomly initialized once from the normalized 3D volume and kept fixed throughout training. For each Anchor, the reference points $\bar{\mathbf{X}}_i$ are established within its local region defined by $r_i$. We then query the Triplane to obtain geometry features as $\mathbf{F}^{\mathrm{geo}}_i=\phi(\bar{\mathbf{X}}_i)\in\mathbb{R}^{K\times3C_T}$, where $\phi(\cdot)$ denotes the Triplane sampling function in Eq.~(\ref{eq:sample}). To refine $\bar{\mathbf{X}}_i$ based on $\mathbf{F}^{\mathrm{geo}}_i$, we introduce $K$ learnable tokens $\mathbf{Q}^{\mathrm{off}}$ and apply cross-attention to predict coordinate offsets:
\begin{equation}
\Delta\mathbf{X}_i
=\operatorname{CrossAtt}\left(\mathbf{Q}^{\mathrm{off}}_i,\mathbf{F}^{\mathrm{geo}}_i,
\mathbf{F}^{\mathrm{geo}}_i\right).
\label{eq:point_offset}
\end{equation}
We scale the offsets by $r_i$ and obtain $K$ points $\mathbf{X}_i$ as $\mathbf{X}_i=\bar{\mathbf{X}}_i+r_i\Delta\mathbf{X}_i$, where
$\mathbf{X}_i\in\mathbb{R}^{K\times3}$.
The final dense points are defined as $\mathcal{P}=[\mathbf{X}_1;\ldots;\mathbf{X}_M]\in\mathbb{R}^{N\times3}$ with $N=M\times K$.

\textbf{Point-to-Gaussian Decoding.} To further decode $\mathcal{G}$ from the dense points $\mathcal{P}$, we start by querying $\mathcal{T}$ at each refined point $\mathbf{x}_j\in\mathcal{P}$. A shared MLP is then employed to produce the Gaussian features as $\mathbf{z}_j=\operatorname{MLP}([\mathbf{x}_j;\phi(\mathbf{x}_j)])$. The corresponding $\mathbf{g}_j$ is predicted from $\mathbf{z}_j$ through a Gaussian head as
\begin{equation}
(\Delta\boldsymbol{\mu}_j,\alpha_j,\mathbf{s}_j,\mathbf{q}_j,\mathbf{c}_j)
=\operatorname{Head}_G(\mathbf{z}_j).
\label{eq:gaussian_head}
\end{equation}
The final Gaussian center is computed as $\boldsymbol{\mu}_j=\mathbf{x}_j+\Delta\boldsymbol{\mu}_j$.

\subsection{Training Loss}
\label{sec:training}
The geometric priors provided by the aforementioned Anchors and dense points allow us to enhance the supervision by utilizing ground-truth geometry.
Therefore, we employ a geometry loss defined as the symmetric Chamfer Distance $\operatorname{CD}(\cdot,\cdot)$:
\begin{equation}
\mathcal{L}_{\mathrm{geo}}=\operatorname{CD}(\mathcal{P}_A,\mathcal{P}_{\mathrm{gt}}^{A})
+\operatorname{CD}(\mathcal{P},\mathcal{P}_{\mathrm{gt}}),
\label{eq:geometry_loss}
\end{equation}
where $\mathcal{P}_A$ denotes the Anchor centers, $\mathcal{P}$ indicates the decoded dense 3D points, and $(\mathcal{P}_{\mathrm{gt}}^{A},\mathcal{P}_{\mathrm{gt}})$ are the ground-truth counterparts. We unproject and fuse the valid depth points from the input views. We then apply FPS~\citep{qi2017pointnet++} to obtain $\mathcal{P}_{\mathrm{gt}}\in\mathbb{R}^{N\times3}$ and $\mathcal{P}_{\mathrm{gt}}^{A}\in\mathbb{R}^{M\times3}$, matching the predicted points $\mathcal{P}$ and Anchor centers $\mathcal{P}_A$, respectively. This geometry supervision enables reliable scene-wide Anchor placement and local point refinement.

Furthermore, we render the predicted Gaussians at the input cameras and compute the standard photometric loss~\citep{wang2004image,zhang2018unreasonable} as
\begin{equation}
\mathcal{L}_{\mathrm{render}}=\mathcal{L}_{\mathrm{rgb}}
+\lambda_s\mathcal{L}_{\mathrm{SSIM}}
+\lambda_p\mathcal{L}_{\mathrm{LPIPS}}.
\label{eq:render_loss}
\end{equation}
The complete training loss is defined as $\mathcal{L}=\mathcal{L}_{\mathrm{geo}}+\mathcal{L}_{\mathrm{render}}$.
In practice, we first train with geometric supervision alone for 15K iterations, and then add the rendering loss. This allows the model to establish stable geometry before jointly optimizing geometry and appearance.

\section{Experiments}
\subsection{Experimental Setup}

\textbf{Implementation details.}
The training resolution is set to $256\times256$. The image encoder contains 8 ViT blocks, and the shared attention module contains 12 attention blocks. We use $M=4096$ Anchor tokens and predict $K=16$ dense points per Anchor, yielding a fixed set of $65{,}536$ Gaussians. Triplane is progressively upsampled from resolution $64$ to $128$ and then $256$. We train for 320K iterations using AdamW~\citep{loshchilov2017decoupled}, with a peak learning rate of $4\times10^{-4}$ and BF16 precision. Training is performed on 128 NVIDIA L20 GPUs with a global batch size of 128. During training, the network takes 4 input images, and the renderings are produced at the input cameras, without sampling additional target views.

\textbf{Datasets.}
We train GeoGSplat on five public datasets: ScanNet++~\citep{yeshwanth2023scannet++}, ARKitScenes~\citep{dehghan2021arkitscenes}, BlendedMVS~\citep{yao2020blendedmvs}, CO3D~\citep{reizenstein2021common}, and WildRGB-D~\citep{xia2024rgbd}. We perform evaluations on DTU~\citep{jensen2014large} and NRGBD~\citep{azinovic2022neural} for object- and scene-level reconstruction, respectively. For novel view synthesis, each evaluation sample contains 4 novel views. We use 4 context views by default and also report results with 2 and 12 input views in the Appendix~\ref{app:viewpoint_extrapolation}.

\textbf{Evaluation protocol.}
We evaluate NVS using PSNR, SSIM~\citep{wang2004image}, and LPIPS~\citep{zhang2018unreasonable}. The input and target resolutions are set to $256\times256$ for all methods except AnySplat, whose input images are resized to $448\times448$ to match its input resolution requirement. 3D reconstruction is evaluated using Chamfer distances~\citep{tenenbaum1977parametric} from predicted points to ground truth (Pred$\rightarrow$GT) and from ground truth to predicted points (GT$\rightarrow$Pred), as well as their average, denoted as CD. For GeoGSplat, these metrics are computed on the predicted dense points, while for the other methods, they are computed on the predicted Gaussian centers. 
Notably, the results of GeoGSplat computed on dense points and Gaussian centers are nearly identical, as reported in Appendix~\ref{app:chamfer_distance_comparison}. For a fair comparison, we define ground-truth point clouds in a normalized coordinate system. We then transform predicted point clouds from the camera coordinate system to the same coordinate system for CD computation. All evaluated methods use ground-truth camera poses for the input images, except AnySplat, which uses predicted poses by default. Notably, we also report results of AnySplat with ground-truth poses, denoted as AnySplat$^{*}$.

\subsection{Novel View Synthesis and Geometric Reconstruction}
\label{sec:main_results}

\begin{figure}[!t]
    \centering
    \includegraphics[width=\linewidth]{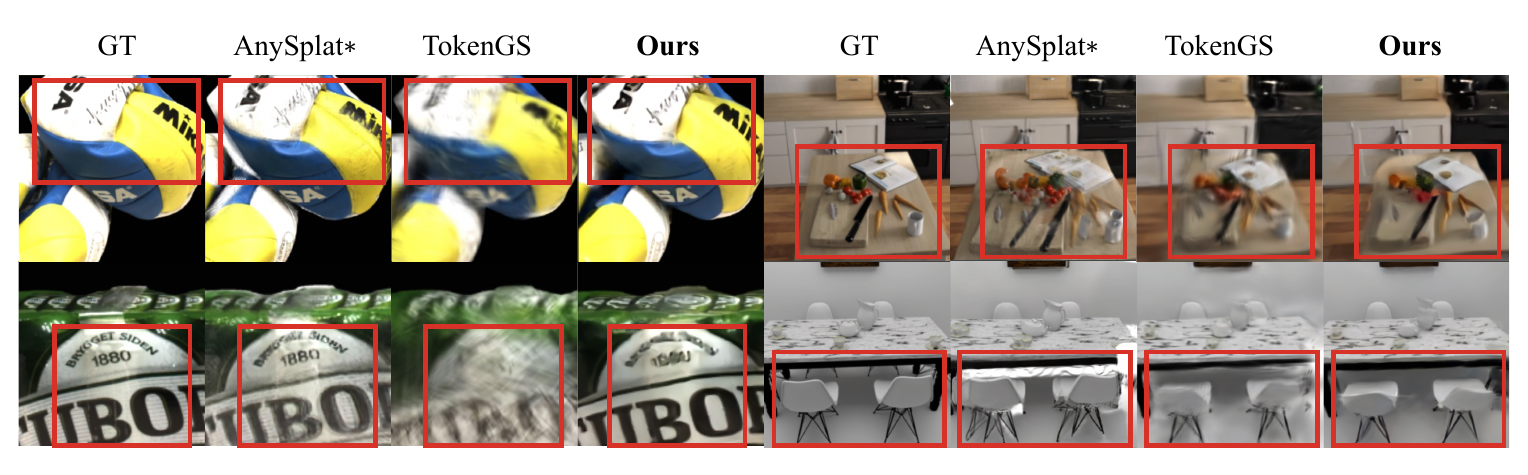}
    \caption{\textbf{Qualitative novel view synthesis with four input views.} Columns show ground truth, AnySplat$^{*}$, TokenGS, and GeoGSplat from left to right. Red boxes highlight blurred details in TokenGS and shifted or duplicated structures in AnySplat. In contrast, GeoGSplat preserves finer details and more accurate scene structures in these regions.}
    \label{fig:qualitative}
    \vspace{-3mm}
\end{figure}

Tables~\ref{tab:main_nvs} and~\ref{tab:main_geometry} report NVS and reconstruction results with 4 input views, and Fig.~\ref{fig:qualitative} provides qualitative comparisons. We compare our GeoGSplat against pixel-aligned methods MVSplat~\citep{chen2024mvsplat}, DepthSplat~\citep{xu2025depthsplat}, and AnySplat~\citep{jiang2025anysplat}. Although AnySplat aggregates Gaussians through voxelization, it relies on per-pixel depth prediction, and the Gaussian count is still coupled to both image resolution and the number of input views. We also compare against non-pixel-aligned methods TokenGS~\citep{tokengs2026} and VolSplat~\citep{wang2026volsplat}.

\textbf{Novel view synthesis.}
Under the same Gaussian budget, GeoGSplat substantially improves NVS quality over TokenGS on both datasets. On DTU, GeoGSplat improves PSNR from 17.71 to 20.48~dB and reduces LPIPS from 0.354 to 0.137. On NRGBD, PSNR improves from 23.82 to 24.78~dB and LPIPS decreases from 0.173 to 0.120. Notably, even compared with pixel-aligned approaches that predict considerably more Gaussians, GeoGSplat still achieves the best PSNR, SSIM, and LPIPS. For example, AnySplat$^{*}$ uses 624K Gaussians on NRGBD, nearly $10\times$ more than GeoGSplat, yet attains lower NVS quality.

\begin{table}[!t]
\caption{\textbf{Novel view synthesis with four input views.} NUM denotes the number of Gaussians used for rendering. The best PSNR, SSIM, and LPIPS are shown in bold.}
\label{tab:main_nvs}
\centering
{\footnotesize
\renewcommand{\arraystretch}{1.08}
\begin{tabular*}{0.9\textwidth}{@{\extracolsep{\fill}}lcccccccc@{}}
\toprule
\multicolumn{1}{c}{\tworowhead{Method}} & \multicolumn{4}{c}{DTU} & \multicolumn{4}{c}{NRGBD} \\
\cmidrule(lr){2-5}\cmidrule(lr){6-9}
& PSNR$\uparrow$ & SSIM$\uparrow$ & LPIPS$\downarrow$ & NUM
& PSNR$\uparrow$ & SSIM$\uparrow$ & LPIPS$\downarrow$ & NUM \\
\midrule
MVSplat & 14.44 & 0.520 & 0.401 & 158K & 18.53 & 0.609 & 0.319 & 255K \\
DepthSplat & 15.46 & 0.599 & 0.331 & 158K & 22.98 & 0.792 & 0.170 & 255K \\
AnySplat & 16.60 & 0.630 & 0.166 & 170K & 20.12 & 0.690 & 0.138 & 643K \\
AnySplat$^{*}$ & 17.54 & 0.700 & 0.148 & 166K & 20.52 & 0.756 & 0.140 & 624K \\
\addlinespace[1pt]
VolSplat & 15.86 & 0.602 & 0.306 & 152K & 22.22 & 0.757 & 0.216 & 246K \\
TokenGS & 17.71 & 0.643 & 0.354 & 65K & 23.82 & 0.844 & 0.173 & 65K \\
\textbf{Ours} & \textbf{20.48} & \textbf{0.771} & \textbf{0.137} & 65K
& \textbf{24.78} & \textbf{0.847} & \textbf{0.120} & 65K \\
\bottomrule
\end{tabular*}}
\end{table}

\begin{table}[!htbp]
\caption{\textbf{3D reconstruction with four input views.} Pred$\rightarrow$GT and GT$\rightarrow$Pred denote Chamfer distances from predicted points to ground-truth points and vice versa, respectively. CD indicates their average. The best results are shown in bold.}
\label{tab:main_geometry}
\centering
{\footnotesize
\renewcommand{\arraystretch}{1.08}
\begin{tabular*}{0.9\textwidth}{@{\extracolsep{\fill}}lcccccc@{}}
\toprule
\multicolumn{1}{c}{\tworowhead{Method}} & \multicolumn{3}{c}{DTU} & \multicolumn{3}{c}{NRGBD} \\
\cmidrule(lr){2-4}\cmidrule(lr){5-7}
& Pred$\rightarrow$GT$\downarrow$ & GT$\rightarrow$Pred$\downarrow$ & CD$\downarrow$
& Pred$\rightarrow$GT$\downarrow$ & GT$\rightarrow$Pred$\downarrow$ & CD$\downarrow$ \\
\midrule
MVSplat & 0.271 & 0.061 & 0.166 & 0.091 & 0.080 & 0.085 \\
DepthSplat & 0.130 & 0.066 & 0.098 & 0.093 & 0.058 & 0.076 \\
AnySplat & 0.048 & 0.032 & 0.040 & 0.063 & 0.046 & 0.055 \\
AnySplat$^{*}$ & 0.033 & \textbf{0.016} & 0.025 & 0.046 & 0.038 & 0.042 \\
\addlinespace[1pt]
VolSplat & 0.091 & 0.035 & 0.063 & 0.061 & 0.059 & 0.060 \\
TokenGS & 0.108 & 0.057 & 0.082 & 0.515 & 0.039 & 0.277 \\
\textbf{Ours} & \textbf{0.019} & \textbf{0.016} & \textbf{0.018}
& \textbf{0.017} & \textbf{0.017} & \textbf{0.017} \\
\bottomrule
\end{tabular*}}
\end{table}

\textbf{3D reconstruction.}
The NVS improvement is accompanied by significantly more accurate geometric reconstruction. As shown in Table~\ref{tab:main_geometry}, GeoGSplat reduces CD from 0.082 to 0.018 on DTU and from 0.277 to 0.017 on NRGBD compared with TokenGS. It also achieves the lowest CD among all compared methods on both datasets. Notably, TokenGS obtains a Pred$\rightarrow$GT distance of 0.515 despite a GT$\rightarrow$Pred distance of only 0.039 on NRGBD, indicating that many predicted centers are displaced from the ground-truth geometry. GeoGSplat reduces both directional distances to 0.017. Compared with pixel-aligned baselines, such as AnySplat$^{*}$, GeoGSplat also achieves lower CD, demonstrating accurate reconstruction without relying on the expensive pixel-aligned predictions.

\textbf{Qualitative comparison.}
Fig.~\ref{fig:qualitative} provides a visual comparison of the synthesized novel views. TokenGS~\citep{tokengs2026} produces blurred object markings on DTU and blurred textures on NRGBD. AnySplat$^{*}$~\citep{jiang2025anysplat} yields sharper renderings, but suffers from duplicated or distorted structures, particularly on NRGBD, due to inconsistent depth predictions across views. In contrast, GeoGSplat preserves both fine appearance details and structurally faithful scene geometry. Such observations show the effectiveness of our geometry-guided non-pixel-aligned paradigm. Our explicit geometric constraints improve geometry over TokenGS, reducing blurred appearance, and non-pixel-aligned prediction implicitly aligns information across views, avoiding duplicated structures caused by the inconsistent depth.

\subsection{How Geometry Affects Novel View Synthesis}
\label{sec:geometry_nvs}

Considering that inaccurate geometry may still produce plausible renderings from viewpoints close to the input views, we examine whether the impact of geometric errors becomes more apparent under larger viewpoint changes. We keep the context views fixed and compare GeoGSplat and TokenGS along the same target-view trajectories on NRGBD. Please refer to the Appendix~\ref{app:target_context_distance} for more details.

\begin{figure}[!htbp]
\centering
\includegraphics[width=0.8\linewidth]{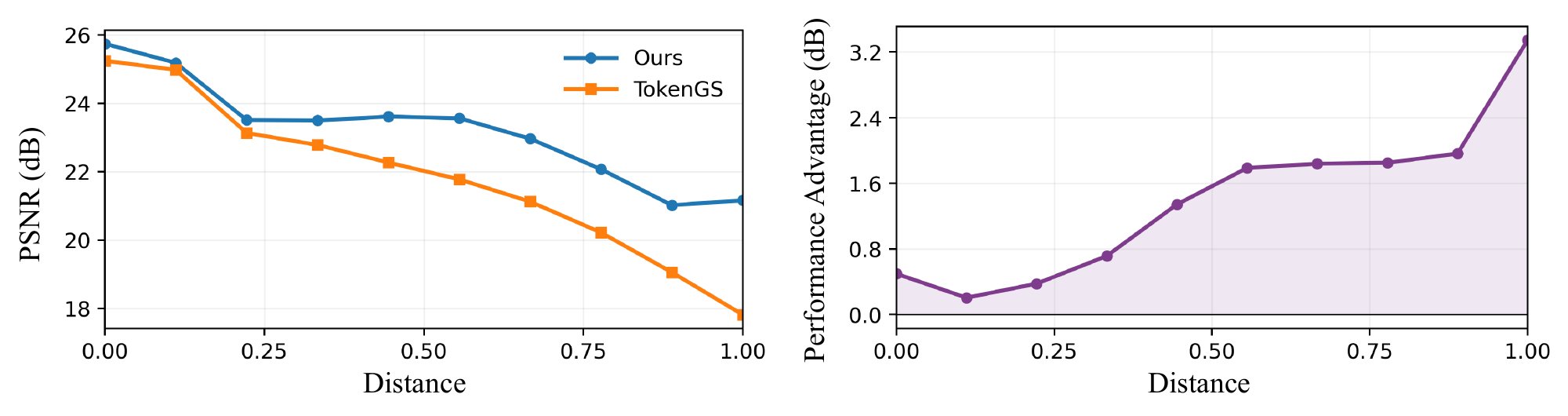}
\caption{\textbf{NVS under varying viewpoint changes on NRGBD.} The plots show PSNR (left) and the performance advantage over TokenGS (right) at different target-context distances. The performance gap between GeoGSplat and TokenGS becomes larger as the target-context distance increases.}
\label{fig:geometry_nvs_distance}
\end{figure}

As shown in Fig.~\ref{fig:geometry_nvs_distance}, both methods achieve similar performance for target views close to the context views. As the distance increases, the performance gap between GeoGSplat and TokenGS becomes larger, with GeoGSplat maintaining higher PSNR. These results suggest that TokenGS can produce plausible renderings near the input views despite inaccurate geometry, but its geometric errors become increasingly apparent as the viewpoint changes. In contrast, GeoGSplat benefits from reliable geometry, leading to more robust NVS under larger viewpoint changes. 
\begin{figure}[!htbp]
    \centering
    \includegraphics[width=\linewidth]{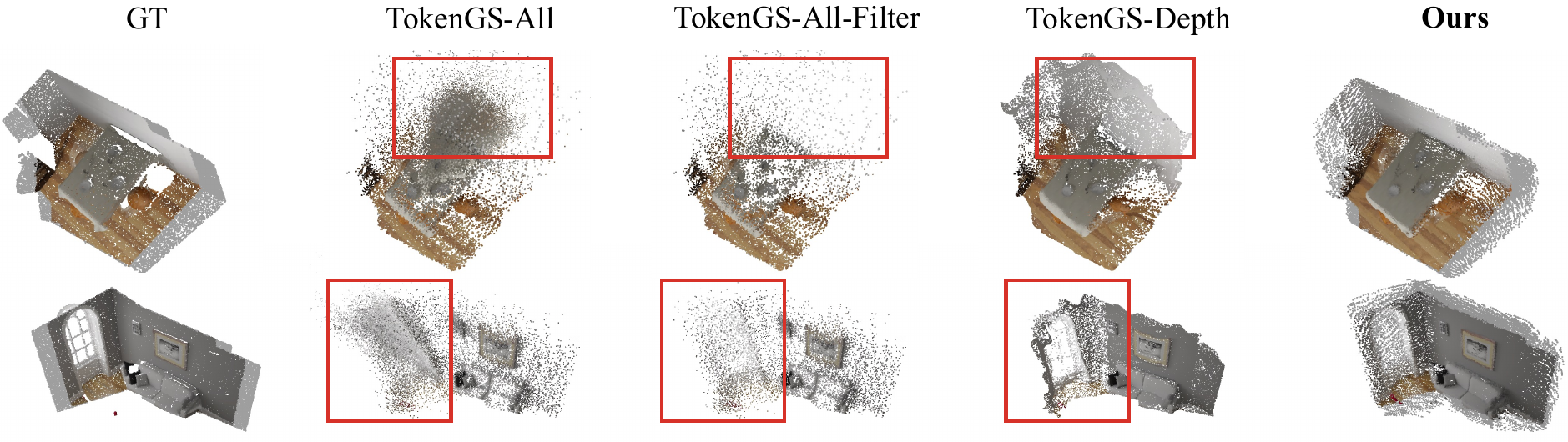}
    \caption{\textbf{3D reconstruction on NRGBD.} From left to right: ground truth, TokenGS Gaussian centers, opacity-filtered TokenGS centers, points unprojected from depth rendered by TokenGS, and GeoGSplat dense points. TokenGS contains substantial geometric outliers that remain after opacity filtering, while its depth-based reconstruction is also displaced from the ground truth. GeoGSplat predicts points closer to the ground-truth geometry and provides more complete scene coverage.}
    \label{fig:geometry_nvs_diagnostics}
\end{figure}
Fig.~\ref{fig:geometry_nvs_diagnostics} provides a detailed comparison of the recovered geometry on NRGBD. TokenGS predicts Gaussian centers that are substantially far from the ground-truth points, consistent with its high Pred$\rightarrow$GT distance reported in Sec.~\ref{sec:main_results}. These outliers remain after removing low-opacity Gaussians, indicating that they are not simply caused by low-contribution primitives. Points unprojected from TokenGS rendered depth also remain displaced from the ground truth. In contrast, GeoGSplat reconstructs points that are much closer to the ground-truth geometry.

\subsection{Ablation Studies}
\label{sec:ablations}

We conduct comprehensive ablations on the key components of GeoGSplat and report the results in Table~\ref{tab:ablations}. All variants use the same data and evaluation protocol, ensuring fair comparisons. For computational efficiency, we conduct all ablation studies using a smaller GeoGSplat model. Please refer to the Appendix~\ref{app:model_architecture} for the detailed model architecture.

\begin{table}[!htbp]
\caption{\textbf{Ablation studies}. Full uses both representations, both decodings, and both geometry losses. The best result in each group is shown in bold.}
\label{tab:ablations}
\centering
{\scriptsize
\renewcommand{\arraystretch}{1.08}
\setlength{\tabcolsep}{2.2pt}
\begin{tabular*}{0.9\textwidth}{@{\extracolsep{\fill}}lcccccccc@{}}
\toprule
\multicolumn{1}{c}{\tworowhead{Variant}} & \multicolumn{4}{c}{DTU} & \multicolumn{4}{c}{NRGBD} \\
\cmidrule(lr){2-5}\cmidrule(lr){6-9}
& PSNR$\uparrow$ & SSIM$\uparrow$ & LPIPS$\downarrow$ & CD$\downarrow$
& PSNR$\uparrow$ & SSIM$\uparrow$ & LPIPS$\downarrow$ & CD$\downarrow$ \\
\midrule
w/o Anchor & 18.67 & 0.700 & 0.188 & 0.031 & 22.46 & 0.782 & 0.185 & 0.030 \\
w/o Triplane & 18.63 & 0.663 & 0.253 & 0.031 & 22.47 & 0.733 & 0.229 & 0.031 \\
\textbf{Full} & \textbf{19.79} & \textbf{0.728} & \textbf{0.183} & \textbf{0.019}
& \textbf{23.70} & \textbf{0.798} & \textbf{0.164} & \textbf{0.019} \\
\midrule
w/o Anchor-to-Point Decoding & 18.32 & 0.653 & 0.276 & 0.032 & 21.81 & 0.703 & 0.276 & 0.041 \\
w/o Point-to-Gaussian Decoding & 18.71 & 0.720 & 0.186 & 0.031 & 22.35 & 0.768 & 0.189 & 0.031 \\
\textbf{Full} & \textbf{19.79} & \textbf{0.728} & \textbf{0.183} & \textbf{0.019}
& \textbf{23.70} & \textbf{0.798} & \textbf{0.164} & \textbf{0.019} \\
\midrule
w/o Geometry Losses & 13.60 & 0.518 & 0.343 & 0.218 & 18.53 & 0.600 & 0.301 & 0.135 \\
w/o Dense-Point Loss & 18.76 & 0.705 & 0.187 & 0.026 & 23.42 & 0.791 & 0.166 & 0.025 \\
w/o Anchor Loss & 19.60 & 0.724 & \textbf{0.178} & 0.020 & \textbf{23.82} & \textbf{0.801} & \textbf{0.161} & 0.022 \\
\textbf{Full} & \textbf{19.79} & \textbf{0.728} & 0.183 & \textbf{0.019}
& 23.70 & 0.798 & 0.164 & \textbf{0.019} \\
\bottomrule
\end{tabular*}}
\end{table}

\textbf{Dual-Branch Structured Encoder.} We first investigate the effectiveness of the dual-branch encoder. w/o Anchor and w/o Triplane remove the Anchor and Triplane constructions, respectively, and directly predict points and Gaussians from $\mathbf{F}^{A}$ and $\mathbf{F}^{T}$. The two variants incur noticeable performance drops on DTU and NRGBD. In particular, PSNR drops by more than 1~dB on both datasets. These results validate the complementary roles of Anchors and Triplane in our encoder.

\textbf{Geometry-Guided Hierarchical Decoder.} We next examine the progressive decoding process. w/o Anchor-to-Point Decoding directly regresses dense points from $\mathbf{F}^{A}$ without querying Triplane. w/o Point-to-Gaussian Decoding removes the query process at the Gaussian prediction stage and predicts Gaussians from $\mathbf{F}^{A}$. Both variants perform worse than \emph{Full}, demonstrating the benefit of progressively refining predictions through the two-stage Triplane querying mechanism.

\textbf{Geometry Supervision.} Since our framework is grounded in explicit geometric priors, we further investigate the effect of geometric supervision. Removing the geometry losses causes severe performance drops, demonstrating the importance of geometric supervision for maintaining reliable geometric priors in GeoGSplat. Removing either loss also degrades performance, with the dense-point loss having a larger impact on both NVS and geometry. \emph{Full} achieves the lowest CD, showing that supervising both Anchor centers and dense points contributes to accurate geometric reconstruction.

\section{Conclusion}

We presented GeoGSplat, a geometry-grounded non-pixel-aligned framework for feed-forward Gaussian Splatting that maintains a fixed Gaussian budget while incorporating explicit geometric priors. GeoGSplat combines a dual-branch structured encoder, which represents the scene through complementary Anchors and Triplane, with a geometry-guided hierarchical decoder that progressively reconstructs dense 3D points before predicting Gaussian primitives. Experiments on DTU and NRGBD demonstrate consistent improvements over both pixel-aligned and non-pixel-aligned baselines in NVS and 3D reconstruction, with the advantage becoming more significant under large viewpoint changes. These results highlight the importance of explicit geometric grounding for robust and generalizable feed-forward Gaussian Splatting.

\section*{AI use statement}

In this work, we used generative AI tools for code debugging, programming assistance, and translation assistance. We have not used generative AI tools for generating synthetic data sets, developing theoretical models or conceptual frameworks, formulating mathematical claims, providing critical ingredients for proving mathematical claims, assisting in the writing of proofs, proposing or refining hypotheses, designing or providing feedback on research methodology or experiments, cleaning or reformatting datasets, supporting qualitative or thematic data analysis, or interpreting results. Additionally, we used generative AI tools for editing the manuscript to improve readability. We have reviewed all AI-assisted work. The AI-assisted code was tested and verified by the authors for correctness and reproducibility, and the AI-assisted text edits were reviewed and revised by the authors to ensure accuracy and clarity. Generative AI was not used for ideation, hypothesis generation, or research design. We take responsibility for the final content of this work, including text, claims, or artifacts produced with the aid of generative AI.

\bibliography{iclr2027_conference}

@article{kerbl20233d,
  title={3d gaussian splatting for real-time radiance field rendering.},
  author={Kerbl, Bernhard and Kopanas, Georgios and Leimk{\"u}hler, Thomas and Drettakis, George and others},
  journal={ACM Transactions on Graphics},
  volume={42},
  number={4},
  pages={139--1},
  year={2023}
}

@inproceedings{chan2022efficient,
  title={Efficient geometry-aware 3d generative adversarial networks},
  author={Chan, Eric R and Lin, Connor Z and Chan, Matthew A and Nagano, Koki and Pan, Boxiao and De Mello, Shalini and Gallo, Orazio and Guibas, Leonidas J and Tremblay, Jonathan and Khamis, Sameh and others},
  booktitle={Proceedings of the IEEE/CVF Conference on Computer Vision and Pattern Recognition},
  pages={16123--16133},
  year={2022}
}

@inproceedings{lu2024scaffold,
  title={Scaffold-gs: Structured 3d gaussians for view-adaptive rendering},
  author={Lu, Tao and Yu, Mulin and Xu, Linning and Xiangli, Yuanbo and Wang, Limin and Lin, Dahua and Dai, Bo},
  booktitle={Proceedings of the IEEE/CVF Conference on Computer Vision and Pattern Recognition},
  pages={20654--20664},
  year={2024}
}

@inproceedings{hong2024lrm,
  title={Lrm: Large reconstruction model for single image to 3d},
  author={Hong, Yicong and Zhang, Kai and Gu, Jiuxiang and Bi, Sai and Zhou, Yang and Liu, Difan and Liu, Feng and Sunkavalli, Kalyan and Bui, Trung and Tan, Hao},
  booktitle={International Conference on Learning Representations},
  volume={2024},
  pages={50678--50702},
  year={2024}
}

@inproceedings{zhang2024gs,
  title={Gs-lrm: Large reconstruction model for 3d gaussian splatting},
  author={Zhang, Kai and Bi, Sai and Tan, Hao and Xiangli, Yuanbo and Zhao, Nanxuan and Sunkavalli, Kalyan and Xu, Zexiang},
  booktitle={European Conference on Computer Vision},
  pages={1--19},
  year={2024}
}

@article{tokengs2026,
  title={TokenGS: Decoupling 3D Gaussian Prediction from Pixels with Learnable Tokens},
  author={Jiawei Ren and Michal Tyszkiewicz and Jiahui Huang and Zan Gojcic},
  journal={Proceedings of the IEEE/CVF Conference on Computer Vision and Pattern Recognition},
  year={2026}
}

@inproceedings{carion2020end,
  title={End-to-end object detection with transformers},
  author={Carion, Nicolas and Massa, Francisco and Synnaeve, Gabriel and Usunier, Nicolas and Kirillov, Alexander and Zagoruyko, Sergey},
  booktitle={European Conference on Computer Vision},
  pages={213--229},
  year={2020}
}

@inproceedings{chen2024mvsplat,
  title={Mvsplat: Efficient 3d gaussian splatting from sparse multi-view images},
  author={Chen, Yuedong and Xu, Haofei and Zheng, Chuanxia and Zhuang, Bohan and Pollefeys, Marc and Geiger, Andreas and Cham, Tat-Jen and Cai, Jianfei},
  booktitle={European Conference on Computer Vision},
  pages={370--386},
  year={2024}
}

@inproceedings{xu2025depthsplat,
  title={Depthsplat: Connecting gaussian splatting and depth},
  author={Xu, Haofei and Peng, Songyou and Wang, Fangjinhua and Blum, Hermann and Barath, Daniel and Geiger, Andreas and Pollefeys, Marc},
  booktitle={Proceedings of the IEEE/CVF Conference on Computer Vision and Pattern Recognition},
  pages={16453--16463},
  year={2025}
}

@inproceedings{wang2026volsplat,
  title={Volsplat: Rethinking feed-forward 3d gaussian splatting with voxel-aligned prediction},
  author={Wang, Weijie and Chen, Yeqing and Zhang, Zeyu and Liu, Hengyu and Wang, Haoxiao and Feng, Zhiyuan and Qin, Wenkang and Chen, Feng and Bian, Jia-Wang and Zhu, Zheng and others},
  booktitle={European Conference on Computer Vision},
  pages={1--19},
  year={2026}
}

@article{jiang2025anysplat,
  title={Anysplat: Feed-forward 3d gaussian splatting from unconstrained views},
  author={Jiang, Lihan and Mao, Yucheng and Xu, Linning and Lu, Tao and Ren, Kerui and Jin, Yichen and Xu, Xudong and Yu, Mulin and Pang, Jiangmiao and Zhao, Feng and others},
  journal={ACM Transactions on Graphics},
  volume={44},
  number={6},
  pages={1--16},
  year={2025},
  publisher={ACM New York, NY, USA}
}

@inproceedings{jensen2014large,
  title={Large scale multi-view stereopsis evaluation},
  author={Jensen, Rasmus and Dahl, Anders and Vogiatzis, George and Tola, Engin and Aan{\ae}s, Henrik},
  booktitle={Proceedings of the IEEE/CVF Conference on Computer Vision and Pattern Recognition},
  pages={406--413},
  year={2014}
}

@inproceedings{azinovic2022neural,
  title={Neural rgb-d surface reconstruction},
  author={Azinovi{\'c}, Dejan and Martin-Brualla, Ricardo and Goldman, Dan B and Nie{\ss}ner, Matthias and Thies, Justus},
  booktitle={Proceedings of the IEEE/CVF Conference on Computer Vision and Pattern Recognition},
  pages={6280--6291},
  year={2022}
}

@inproceedings{yeshwanth2023scannet++,
  title={Scannet++: A high-fidelity dataset of 3d indoor scenes},
  author={Yeshwanth, Chandan and Liu, Yueh-Cheng and Nie{\ss}ner, Matthias and Dai, Angela},
  booktitle={Proceedings of the IEEE/CVF International Conference on Computer Vision},
  pages={12--22},
  year={2023}
}

@article{dehghan2021arkitscenes,
  title={ARKitScenes: A Diverse Real-World Dataset For 3D Indoor Scene Understanding Using Mobile RGB-D Data.},
  author={Dehghan, Afshin and Baruch, Gilad and Chen, Zhuoyuan and Feigin, Yuri and Fu, Peter and Gebauer, Thomas and Kurz, Daniel and Dimry, Tal and Joffe, Brandon and Schwartz, Arik and others},
  journal={NeurIPS Datasets and Benchmarks},
  volume={2},
  number={6},
  pages={16},
  year={2021}
}

@inproceedings{yao2020blendedmvs,
  title={Blendedmvs: A large-scale dataset for generalized multi-view stereo networks},
  author={Yao, Yao and Luo, Zixin and Li, Shiwei and Zhang, Jingyang and Ren, Yufan and Zhou, Lei and Fang, Tian and Quan, Long},
  booktitle={Proceedings of the IEEE/CVF Conference on Computer Vision and Pattern Recognition},
  pages={1787--1796},
  year={2020}
}

@inproceedings{reizenstein2021common,
  title={Common objects in 3d: Large-scale learning and evaluation of real-life 3d category reconstruction},
  author={Reizenstein, Jeremy and Shapovalov, Roman and Henzler, Philipp and Sbordone, Luca and Labatut, Patrick and Novotny, David},
  booktitle={Proceedings of the IEEE/CVF International Conference on Computer Vision},
  pages={10881--10891},
  year={2021}
}

@inproceedings{xia2024rgbd,
  title={Rgbd objects in the wild: Scaling real-world 3d object learning from rgb-d videos},
  author={Xia, Hongchi and Fu, Yang and Liu, Sifei and Wang, Xiaolong},
  booktitle={Proceedings of the IEEE/CVF Conference on Computer Vision and Pattern Recognition},
  pages={22378--22389},
  year={2024}
}

@article{qi2017pointnet++,
  title={Pointnet++: Deep hierarchical feature learning on point sets in a metric space},
  author={Qi, Charles Ruizhongtai and Yi, Li and Su, Hao and Guibas, Leonidas J},
  journal={Advances in neural information processing systems},
  volume={30},
  year={2017}
}

@article{wang2004image,
  title={Image quality assessment: from error visibility to structural similarity},
  author={Wang, Zhou and Bovik, Alan C and Sheikh, Hamid R and Simoncelli, Eero P},
  journal={IEEE Transactions on Image Processing},
  volume={13},
  number={4},
  pages={600--612},
  year={2004},
  publisher={IEEE}
}

@inproceedings{zhang2018unreasonable,
  title={The unreasonable effectiveness of deep features as a perceptual metric},
  author={Zhang, Richard and Isola, Phillip and Efros, Alexei A and Shechtman, Eli and Wang, Oliver},
  booktitle={Proceedings of the IEEE/CVF Conference on Computer Vision and Pattern Recognition},
  pages={586--595},
  year={2018}
}

@article{loshchilov2017decoupled,
  title={Decoupled weight decay regularization},
  author={Loshchilov, Ilya and Hutter, Frank},
  journal={arXiv preprint arXiv:1711.05101},
  year={2017}
}

@inproceedings{wang2025vggt,
  title={Vggt: Visual geometry grounded transformer},
  author={Wang, Jianyuan and Chen, Minghao and Karaev, Nikita and Vedaldi, Andrea and Rupprecht, Christian and Novotny, David},
  booktitle={Proceedings of the IEEE/CVF Conference on Computer Vision and Pattern Recognition},
  pages={5294--5306},
  year={2025}
}

@inproceedings{he2016deep,
  title={Deep residual learning for image recognition},
  author={He, Kaiming and Zhang, Xiangyu and Ren, Shaoqing and Sun, Jian},
  booktitle={Proceedings of the IEEE/CVF Conference on Computer Vision and Pattern Recognition},
  pages={770--778},
  year={2016}
}

@inproceedings{wang2023rodin,
  title={Rodin: A generative model for sculpting 3d digital avatars using diffusion},
  author={Wang, Tengfei and Zhang, Bo and Zhang, Ting and Gu, Shuyang and Bao, Jianmin and Baltrusaitis, Tadas and Shen, Jingjing and Chen, Dong and Wen, Fang and Chen, Qifeng and others},
  booktitle={Proceedings of the IEEE/CVF Conference on Computer Vision and Pattern Recognition},
  pages={4563--4573},
  year={2023}
}

@inproceedings{charatan2024pixelsplat,
  title={pixelsplat: 3d gaussian splats from image pairs for scalable generalizable 3d reconstruction},
  author={Charatan, David and Li, Sizhe Lester and Tagliasacchi, Andrea and Sitzmann, Vincent},
  booktitle={Proceedings of the IEEE/CVF Conference on Computer Vision and Pattern Recognition},
  pages={19457--19467},
  year={2024}
}

@article{zhang2026anchorsplat,
  title={AnchorSplat: Feed-Forward 3D Gaussian Splatting with 3D Geometric Priors},
  author={Zhang, Xiaoxue and Zheng, Xiaoxu and Yin, Yixuan and Zhao, Tiao and Tang, Kaihua and Mi, Michael Bi and Xu, Zhan and Chen, Dave Zhenyu},
  journal={arXiv preprint arXiv:2604.07053},
  year={2026}
}

@inproceedings{yao2018mvsnet,
  title={Mvsnet: Depth inference for unstructured multi-view stereo},
  author={Yao, Yao and Luo, Zixin and Li, Shiwei and Fang, Tian and Quan, Long},
  booktitle={European Conference on Computer Vision},
  pages={785--801},
  year={2018}
}

@inproceedings{gu2020cascade,
  title={Cascade cost volume for high-resolution multi-view stereo and stereo matching},
  author={Gu, Xiaodong and Fan, Zhiwen and Zhu, Siyu and Dai, Zuozhuo and Tan, Feitong and Tan, Ping},
  booktitle={Proceedings of the IEEE/CVF Conference on Computer Vision and Pattern Recognition},
  pages={2492--2501},
  year={2020}
}

@inproceedings{yu2021pixelnerf,
  title={pixelnerf: Neural radiance fields from one or few images},
  author={Yu, Alex and Ye, Vickie and Tancik, Matthew and Kanazawa, Angjoo},
  booktitle={Proceedings of the IEEE/CVF Conference on Computer Vision and Pattern Recognition},
  pages={4576--4585},
  year={2021}
}

@inproceedings{wang2021ibrnet,
  title={Ibrnet: Learning multi-view image-based rendering},
  author={Wang, Qianqian and Wang, Zhicheng and Genova, Kyle and Srinivasan, Pratul P and Zhou, Howard and Barron, Jonathan T and Martin-Brualla, Ricardo and Snavely, Noah and Funkhouser, Thomas},
  booktitle={Proceedings of the IEEE/CVF Conference on Computer Vision and Pattern Recognition},
  pages={4690--4699},
  year={2021}
}

@inproceedings{chen2021mvsnerf,
  title={Mvsnerf: Fast generalizable radiance field reconstruction from multi-view stereo},
  author={Chen, Anpei and Xu, Zexiang and Zhao, Fuqiang and Zhang, Xiaoshuai and Xiang, Fanbo and Yu, Jingyi and Su, Hao},
  booktitle={Proceedings of the IEEE/CVF International Conference on Computer Vision},
  pages={14104--14113},
  year={2021}
}

@inproceedings{johari2022geonerf,
  title={Geonerf: Generalizing nerf with geometry priors},
  author={Johari, Mohammad Mahdi and Lepoittevin, Yann and Fleuret, Fran{\c{c}}ois},
  booktitle={Proceedings of the IEEE/CVF Conference on Computer Vision and Pattern Recognition},
  pages={18344--18347},
  year={2022}
}

@inproceedings{wang2024dust3r,
  title={Dust3r: Geometric 3d vision made easy},
  author={Wang, Shuzhe and Leroy, Vincent and Cabon, Yohann and Chidlovskii, Boris and Revaud, Jerome},
  booktitle={Proceedings of the IEEE/CVF Conference on Computer Vision and Pattern Recognition},
  pages={20697--20709},
  year={2024}
}

@inproceedings{leroy2024grounding,
  title={Grounding image matching in 3d with mast3r},
  author={Leroy, Vincent and Cabon, Yohann and Revaud, J{\'e}r{\^o}me},
  booktitle={European Conference on Computer Vision},
  pages={71--91},
  year={2024}
}

@inproceedings{szymanowicz2024splatter,
  title={Splatter image: Ultra-fast single-view 3d reconstruction},
  author={Szymanowicz, Stanislaw and Rupprecht, Chrisitian and Vedaldi, Andrea},
  booktitle={Proceedings of the IEEE/CVF Conference on Computer Vision and Pattern Recognition},
  pages={10208--10217},
  year={2024}
}

@inproceedings{zhao2026rayzer,
  title={E-rayzer: Self-supervised 3d reconstruction as spatial visual pre-training},
  author={Zhao, Qitao and Tan, Hao and Wang, Qianqian and Bi, Sai and Zhang, Kai and Sunkavalli, Kalyan and Tulsiani, Shubham and Jiang, Hanwen},
  booktitle={Proceedings of the IEEE/CVF Conference on Computer Vision and Pattern Recognition},
  pages={7525--7535},
  year={2026}
}

@inproceedings{moreau2026off,
  title={Off the grid: Detection of primitives for feed-forward 3d gaussian splatting},
  author={Moreau, Arthur and Shaw, Richard and Nazarczuk, Michal and Shin, Jisu and Tanay, Thomas and Zhang, Zhensong and Xu, Songcen and P{\'e}rez-Pellitero, Eduardo},
  booktitle={Proceedings of the IEEE/CVF Conference on Computer Vision and Pattern Recognition},
  pages={11756--11766},
  year={2026}
}

@inproceedings{fridovich2022plenoxels,
  title={Plenoxels: Radiance fields without neural networks},
  author={Fridovich-Keil, Sara and Yu, Alex and Tancik, Matthew and Chen, Qinhong and Recht, Benjamin and Kanazawa, Angjoo},
  booktitle={Proceedings of the IEEE/CVF Conference on Computer Vision and Pattern Recognition},
  pages={5491--5500},
  year={2022}
}

@article{dosovitskiy2020vit,
  title={An Image is Worth 16x16 Words: Transformers for Image Recognition at Scale},
  author={Dosovitskiy, Alexey and Beyer, Lucas and Kolesnikov, Alexander and Weissenborn, Dirk and Zhai, Xiaohua and Unterthiner, Thomas and  Dehghani, Mostafa and Minderer, Matthias and Heigold, Georg and Gelly, Sylvain and Uszkoreit, Jakob and Houlsby, Neil},
  journal={International Conference on Learning Representations},
  year={2021}
}

@article{li2026locusgs,
  title={LocusGS: Spatially Grounded Tokens for Feed-Forward 3D Gaussian Splatting},
  author={Li, Wenyu and Liu, Sidun and Hu, Tongrui and Qiao, Peng and Dou, Yong},
  journal={arXiv preprint arXiv:2608.12825},
  year={2026}
}

@inproceedings{tenenbaum1977parametric,
  title={Parametric correspondence and chamfer matching: Two new techniques for image matching},
  author={Tenenbaum, RC Bolles HG Barrow JM and Bolles, RC and Wolf, HC and others},
  booktitle={Proceedings of the International Joint Conference on Artificial Intelligence},
  pages={659--663},
  year={1977}
}

@article{li2026querysplat,
  title={QuerySplat: Decoupling Geometry and Appearance Representations in 3DGS Prediction},
  author={Li, Yinglong and Shen, Donghui and Zhang, Xiaoyu and Ye, Zhichao and Wu, Hongyu and Hao, Aimin and Zhang, Guofeng and Liu, Haomin},
  journal={arXiv preprint arXiv:2608.01186},
  year={2026}
}

@article{plucker1865new,
  title={I. on a new geometry of space},
  author={Plucker, Julius},
  journal={Proceedings of the Royal Society of London},
  pages={53--58},
  year={1865},
  publisher={The Royal Society London}
}

@inproceedings{jiang2025rayzer,
  title={Rayzer: A self-supervised large view synthesis model},
  author={Jiang, Hanwen and Tan, Hao and Wang, Peng and Jin, Haian and Zhao, Yue and Bi, Sai and Zhang, Kai and Luan, Fujun and Sunkavalli, Kalyan and Huang, Qixing and others},
  booktitle={Proceedings of the IEEE/CVF International Conference on Computer Vision},
  pages={4918--4929},
  year={2025}
}

@inproceedings{barron2022mip,
  title={Mip-nerf 360: Unbounded anti-aliased neural radiance fields},
  author={Barron, Jonathan T and Mildenhall, Ben and Verbin, Dor and Srinivasan, Pratul P and Hedman, Peter},
  booktitle={Proceedings of the IEEE/CVF Conference on Computer Vision and Pattern Recognition},
  pages={5460--5469},
  year={2022}
}

@inproceedings{xiang2026native,
  title={Native and compact structured latents for 3d generation},
  author={Xiang, Jianfeng and Chen, Xiaoxue and Xu, Sicheng and Wang, Ruicheng and Lv, Zelong and Deng, Yu and Zhu, Hongyuan and Dong, Yue and Zhao, Hao and Yuan, Nicholas Jing and others},
  booktitle={Proceedings of the IEEE/CVF Conference on Computer Vision and Pattern Recognition},
  pages={14419--14429},
  year={2026}
}
\bibliographystyle{iclr2027_conference}
\newpage
\appendix
\section{More Experiments}
\subsection{Comparison with Retrained TokenGS}
\label{app:retrained_tokengs}

The evaluation of TokenGS in Sec.~\ref{sec:main_results} uses the publicly released model, which differs from GeoGSplat in training data. To further isolate the effect of our geometry-grounded design, we retrain TokenGS under the same setting as GeoGSplat. Both variants are trained under the same training data, input resolution, Gaussian budget, and training schedule. \emph{TokenGS-R} uses the original novel-view photometric and visibility losses, while \emph{TokenGS-R + Geometry loss} additionally uses the geometry loss on the Gaussian centers. The results are reported in Table~\ref{tab:retrained_tokengs}.

\begin{table}[!htbp]
\caption{\textbf{Comparison with retrained TokenGS under four input views.} Both TokenGS variants use the same setting as GeoGSplat and differ only in whether the geometry loss is applied.}
\label{tab:retrained_tokengs}
\centering
{\footnotesize
\renewcommand{\arraystretch}{1.08}
\setlength{\tabcolsep}{2.2pt}
\begin{tabular*}{\textwidth}{@{\extracolsep{\fill}}lcccccccc@{}}
\toprule
\multicolumn{1}{c}{\tworowhead{Variant}} & \multicolumn{4}{c}{DTU} & \multicolumn{4}{c}{NRGBD} \\
\cmidrule(lr){2-5}\cmidrule(lr){6-9}
& PSNR$\uparrow$ & SSIM$\uparrow$ & LPIPS$\downarrow$ & CD$\downarrow$
& PSNR$\uparrow$ & SSIM$\uparrow$ & LPIPS$\downarrow$ & CD$\downarrow$ \\
\midrule
TokenGS-R & 15.24 & 0.593 & 0.317 & 0.460
& 23.24 & 0.778 & 0.176 & 0.191 \\
TokenGS-R + Geometry loss & 16.37 & 0.608 & 0.314 & 0.038
& 24.43 & 0.807 & 0.153 & 0.029 \\
\textbf{GeoGSplat} & \textbf{20.48} & \textbf{0.771} & \textbf{0.137} & \textbf{0.018}
& \textbf{24.78} & \textbf{0.847} & \textbf{0.120} & \textbf{0.017} \\
\bottomrule
\end{tabular*}}
\end{table}

Both retrained TokenGS variants perform substantially worse than GeoGSplat on both datasets. Adding the geometry loss improves both NVS and geometric accuracy, but a substantial gap remains. Together with the ablations in Table~\ref{tab:ablations}, these results demonstrate that the improvements of GeoGSplat are enabled by the proposed architecture, which is closely integrated with geometry supervision, rather than by geometry supervision alone.

\subsection{Generalization Across Input View Counts}
\label{app:viewpoint_extrapolation}

We extend the evaluation by testing the methods on 2 and 12 input views without retraining, and report the results in Table~\ref{tab:viewpoint_extrapolation}. With 2 input views, GeoGSplat achieves the highest PSNR among the compared methods on both datasets, and the lowest CD on NRGBD. With 12 input views, GeoGSplat maintains its leading performance, further improving PSNR to 22.53~dB on DTU and 25.21~dB on NRGBD, and reducing CD to 0.014 and 0.015, respectively. It is worth noting that GeoGSplat keeps a fixed set of 65K Gaussians for both input settings. These results demonstrate that GeoGSplat can generalize across varying numbers of input views without retraining, while maintaining accurate geometry and high-quality NVS.

\begin{table}[!htbp]
\caption{\textbf{Evaluation under varying input view counts on DTU and NRGBD.} All methods are tested with 2 and 12 input views without retraining. We report PSNR (P) for novel view synthesis, Chamfer distance (CD) for 3D reconstruction, and the number of predicted Gaussians (NUM).}
\label{tab:viewpoint_extrapolation}
\centering
\begin{subtable}[t]{0.495\textwidth}
\centering
\caption{DTU}
\label{tab:viewpoint_extrapolation_dtu}
{\footnotesize
\renewcommand{\arraystretch}{1.08}
\setlength{\tabcolsep}{0.6pt}
\begin{tabular*}{\linewidth}{@{\extracolsep{\fill}}lccc@{\hspace{3pt}}ccc@{}}
\toprule
\multicolumn{1}{c}{\tworowhead{Method}} & \multicolumn{3}{c}{2 Views} & \multicolumn{3}{c}{12 Views} \\
\cmidrule(lr){2-4}\cmidrule(lr){5-7}
& P$\uparrow$ & CD$\downarrow$ & NUM & P$\uparrow$ & CD$\downarrow$ & NUM \\
\midrule
DepthSplat & 14.01 & 0.224 & 78K & 17.06 & 0.035 & 480K \\
AnySplat$^{*}$ & 15.94 & \textbf{0.034} & 91K & 18.71 & 0.019 & 352K \\
TokenGS & 14.57 & 0.155 & 65K & 18.30 & 0.073 & 65K \\
\textbf{Ours} & \textbf{17.53} & 0.037 & 65K & \textbf{22.53} & \textbf{0.014} & 65K \\
\bottomrule
\end{tabular*}}
\end{subtable}\hfill
\begin{subtable}[t]{0.495\textwidth}
\centering
\caption{NRGBD}
\label{tab:viewpoint_extrapolation_nrgbd}
{\footnotesize
\renewcommand{\arraystretch}{1.08}
\setlength{\tabcolsep}{0.6pt}
\begin{tabular*}{\linewidth}{@{\extracolsep{\fill}}lccc@{\hspace{3pt}}ccc@{}}
\toprule
\multicolumn{1}{c}{\tworowhead{Method}} & \multicolumn{3}{c}{2 Views} & \multicolumn{3}{c}{12 Views} \\
\cmidrule(lr){2-4}\cmidrule(lr){5-7}
& P$\uparrow$ & CD$\downarrow$ & NUM & P$\uparrow$ & CD$\downarrow$ & NUM \\
\midrule
DepthSplat & 20.16 & 0.102 & 127K & 22.12 & 0.090 & 766K \\
AnySplat$^{*}$ & 17.28 & 0.045 & 325K & 22.37 & 0.035 & 1665K \\
TokenGS & 21.84 & 0.330 & 65K & 23.90 & 0.250 & 65K \\
\textbf{Ours} & \textbf{22.01} & \textbf{0.024} & 65K & \textbf{25.21} & \textbf{0.015} & 65K \\
\bottomrule
\end{tabular*}}
\end{subtable}
\end{table}

\subsection{Experiments on Mip-NeRF 360}
\label{app:exp_on_mipnerf360}
We conducted additional experiments on Mip-NeRF 360~\citep{barron2022mip}, a large-scale scene-level dataset. As shown in Table~\ref{tab:mipnerf360_12v}, GeoGSplat achieves the best PSNR of 21.59~dB and the lowest CD of 0.015 on Mip-NeRF 360, demonstrating the effectiveness of our method on larger scenes under a fixed Gaussian budget.
\begin{table}[!htbp]
\caption{\textbf{Comparison on Mip-NeRF 360 with 12 input views.} The best results are shown in bold.}
\label{tab:mipnerf360_12v}
\centering
{\footnotesize
\renewcommand{\arraystretch}{1.08}
\setlength{\tabcolsep}{8pt}
\begin{tabular*}{0.85\textwidth}{@{\extracolsep{\fill}}lcccccc@{}}
\toprule
& MVSplat & DepthSplat & AnySplat$^{*}$ & VolSplat & TokenGS & \textbf{Ours} \\
\midrule
PSNR$\uparrow$ & 16.24 & 16.06 & 18.65 & 19.79 & 21.51 & \textbf{21.59} \\
CD$\downarrow$ & 0.106 & 0.063 & 0.081 & 0.058 & 0.132 & \textbf{0.015} \\
\bottomrule
\end{tabular*}}
\end{table}

\subsection{Chamfer Distance Comparison between Dense Points and Gaussian Centers}
\label{app:chamfer_distance_comparison}

We compare the Chamfer distance computed on the dense points and the final Gaussian centers after Point-to-Gaussian decoding. As shown in Table~\ref{tab:dense_vs_gaussian}, the two measurements are nearly identical across all settings. The predicted Gaussian offsets adjust Gaussian centers for better rendering quality, while preserving the high-quality geometry reconstructed by the dense points.

\begin{table}[!htbp]
\caption{\textbf{Consistency between dense points and Gaussian centers.}
We report Chamfer distance computed on the dense points from Anchor-to-Point decoding and the final Gaussian centers after Point-to-Gaussian decoding.}
\label{tab:dense_vs_gaussian}
\centering
{\footnotesize
\renewcommand{\arraystretch}{1.08}
\begin{tabular}{lccc}
\toprule
Dataset & Views & Dense Points CD$\downarrow$ & Gaussian Centers CD$\downarrow$ \\
\midrule
DTU & 2V  & 0.037148 & 0.037130 \\
DTU & 4V  & 0.017984 & 0.017879 \\
DTU & 12V & 0.014296 & 0.014189 \\
\midrule
NRGBD & 2V  & 0.023718 & 0.023623 \\
NRGBD & 4V  & 0.017406 & 0.017485 \\
NRGBD & 12V & 0.015325 & 0.015472 \\
\bottomrule
\end{tabular}}
\end{table}

\begin{figure}[!htbp]
    \centering
    \includegraphics[width=\linewidth]{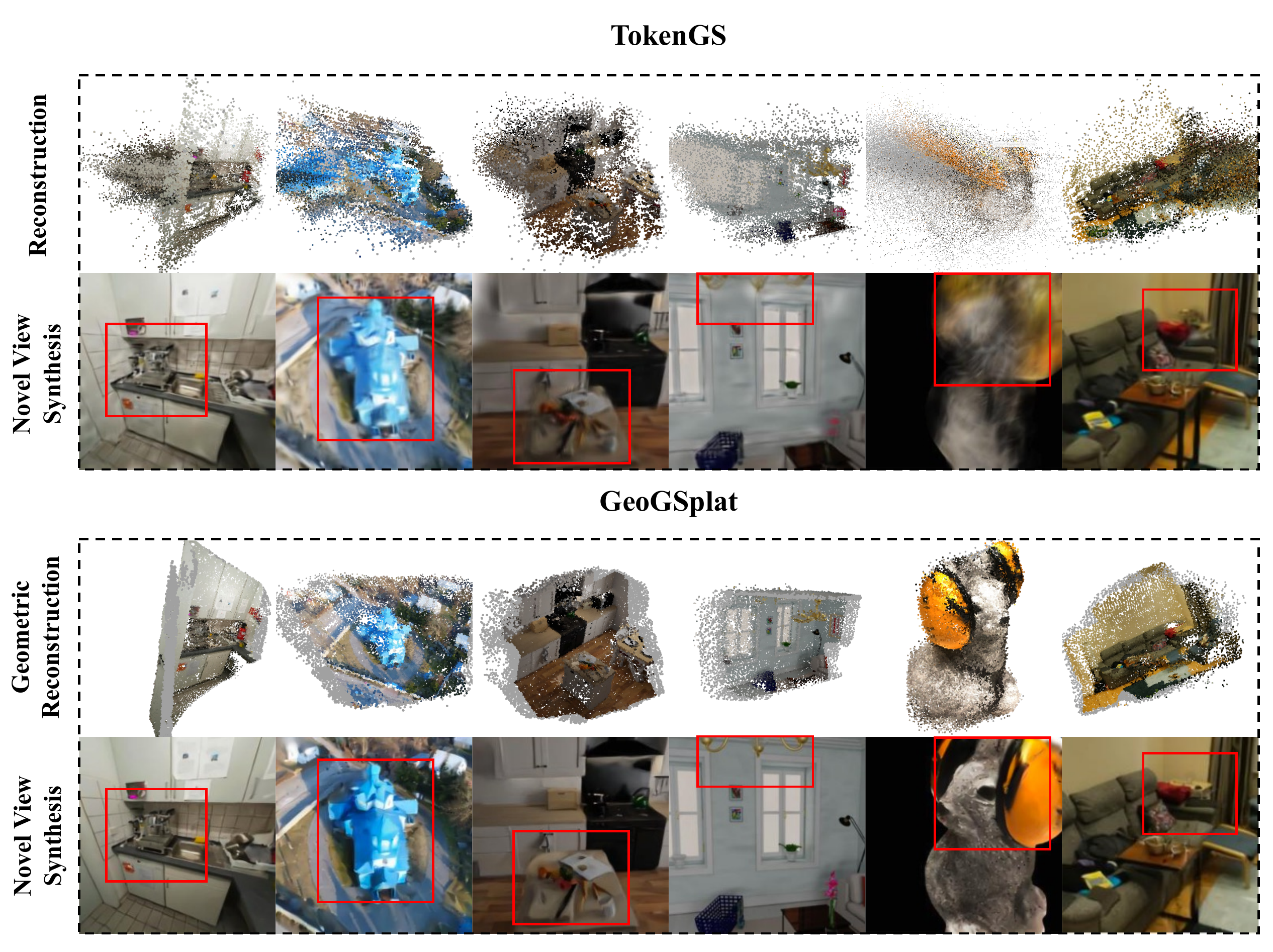}
    \caption{\textbf{Comparison of reconstruction and novel view synthesis.}}
    \label{fig:appendix}
\end{figure}

\subsection{Additional Visualizations}
We provide more visualizations in Fig.~\ref{fig:appendix}, where we can observe that our GeoGSplat reconstructs cleaner Gaussian centers and renders higher-quality images than TokenGS.

\subsection{Representation Efficiency}
\label{app:representation_rfficiency}
The efficiency emphasized in our paper primarily refers to the representation efficiency of non-pixel-aligned methods over pixel-aligned methods. Pixel-aligned methods produce Gaussians in proportion to the input resolution and number of views, leading to substantial redundancy, whereas non-pixel-aligned methods maintain a fixed Gaussian budget. To investigate whether this representation efficiency translates into practical computational benefits, we profile GeoGSplat against AnySplat across 12-64 input views, reporting inference latency, FLOPs, rendering throughput, and peak GPU memory. As shown in Table~\ref{tab:efficiency}, the computational advantages become more pronounced as the number of input views increases.

\begin{table}[!htbp]
\caption{\textbf{Computational efficiency comparison across 12--64 input views.} GeoGSplat maintains a fixed Gaussian budget, with increasingly pronounced computational advantages over AnySplat as the number of input views increases. Parentheses indicate resource usage relative to AnySplat for Gaussian count, inference latency, FLOPs, and peak GPU memory, and relative rendering throughput ($\times$) for FPS.}
\label{tab:efficiency}
\centering
{\footnotesize
\renewcommand{\arraystretch}{1.08}
\setlength{\tabcolsep}{2.2pt}
\begin{tabular*}{\textwidth}{@{\extracolsep{\fill}}clccccc@{}}
\toprule
Views & Method & Gaussians$\downarrow$ & Inference (ms)$\downarrow$ & FLOPs (TF)$\downarrow$ & Rendering FPS$\uparrow$ & Peak Memory (GiB)$\downarrow$ \\
\midrule
12 & \textbf{GeoGSplat} & 65,536 (\textbf{4.0\%}) & 299 (\textbf{71.7\%}) & 38.79 (\textbf{84.5\%}) & 493 (\textbf{1.41$\times$}) & 3.99 (\textbf{65.7\%}) \\
   & AnySplat & 1,651,460 & 417 & 45.90 & 350 & 6.07 \\
\midrule
24 & \textbf{GeoGSplat} & 65,536 (\textbf{2.1\%}) & 443 (\textbf{45.6\%}) & 66.07 (\textbf{54.3\%}) & 619 (\textbf{2.47$\times$}) & 4.14 (\textbf{44.1\%}) \\
   & AnySplat & 3,164,598 & 972 & 121.78 & 251 & 9.38 \\
\midrule
48 & \textbf{GeoGSplat} & 65,536 (\textbf{1.1\%}) & 856 (\textbf{34.3\%}) & 150.31 (\textbf{41.4\%}) & 661 (\textbf{4.11$\times$}) & 4.44 (\textbf{27.8\%}) \\
   & AnySplat & 5,742,420 & 2,494 & 363.47 & 161 & 15.99 \\
\midrule
64 & \textbf{GeoGSplat} & 65,536 (\textbf{0.9\%}) & 1,229 (\textbf{32.4\%}) & 228.46 (\textbf{38.6\%}) & 578 (\textbf{4.22$\times$}) & 4.64 (\textbf{22.8\%}) \\
   & AnySplat & 7,074,148 & 3,791 & 591.21 & 137 & 20.38 \\
\bottomrule
\end{tabular*}}
\vspace{2pt}
\end{table}



\section{Implementation Details}
\label{app:implementation}

\subsection{Shared Attention Module}
\label{app:shared_attention_module}
The attention module $\operatorname{Att}(\cdot)$ used in Eq.~\ref{eq:feature_decoder} consists of a cross-attention layer, a self-attention layer, and a feed-forward network (FFN). Given query tokens $\mathbf{Q}$ and image features $\mathbf{F}_{\mathrm{img}}$, we first perform cross-attention to aggregate image features:
\begin{equation}
\mathbf{H}_1
=
\mathbf{Q}
+
\operatorname{MHA}\left(
\mathbf{Q},
\mathbf{F}_{\mathrm{img}},
\mathbf{F}_{\mathrm{img}}
\right),
\label{eq:att_cross}
\end{equation}
where $\operatorname{MHA}(\cdot,\cdot,\cdot)$ denotes multi-head attention. We then apply self-attention among the query tokens:
\begin{equation}
\mathbf{H}_2
=
\mathbf{H}_1
+
\operatorname{MHA}\left(
\mathbf{H}_1,
\mathbf{H}_1,
\mathbf{H}_1
\right).
\label{eq:att_self}
\end{equation}
The output is further refined by an FFN:
\begin{equation}
\mathbf{H}_3
=
\mathbf{H}_2
+
\operatorname{FFN}\left(\mathbf{H}_2\right),
\label{eq:att_ffn}
\end{equation}
where $\operatorname{FFN}(\cdot)$ consists of two linear layers. The same attention module is shared between the Anchor and Triplane branches, while the two branches use independent query tokens.

\subsection{Triplane Refinement Block}
\label{app:architecture}

Each refinement stage in Eq.~(\ref{eq:triplane_refine}) consists of three operations: a residual convolution that refines each plane locally, a cross-plane fusion that exchanges information between the three planes, and an image-conditioned cross-attention that re-injects image evidence at the new resolution.

First, the Triplane resolution is doubled, followed by two residual convolutional blocks:
\begin{equation}
\mathbf{Z}^{a}
\leftarrow
\mathbf{Z}^{a}
+
\operatorname{Conv}_{3\times3}
\left(
\operatorname{Conv}_{3\times3}(\mathbf{Z}^{a})
\right),
\end{equation}
where $\mathbf{Z}^{a}$ denotes the feature map of plane $a\in\{xy,xz,yz\}$, and $\operatorname{Conv}_{3\times3}$ denotes a $3\times3$ convolution. We then perform cross-plane fusion by exchanging features between the three orthogonal planes. For instance, the $xy$ and $xz$ planes share the $x$-axis. We average the features of the $xz$ plane along the $z$-axis and replicate the resulting features along the $y$-axis to match the $xy$ plane:
\begin{equation}
\mathbf{P}^{xz\rightarrow xy}
=
\operatorname{Expand}_{y}
\left(
\operatorname{Mean}_{z}(\mathbf{Z}^{xz})
\right).
\end{equation}
where $\operatorname{Mean}_{z}$ averages along the $z$-axis, and $\operatorname{Expand}_{y}$ replicates the resulting features along the $y$-axis. The same operation is applied to the other pairs of orthogonal planes. The pooled features from the other two planes are concatenated with the target plane and fused by convolution:
\begin{equation}
\mathbf{Z}^{a}
\leftarrow
\mathbf{Z}^{a}
+
\operatorname{Conv}
\left(
\left[
\mathbf{Z}^{a};
\mathbf{P}^{b\rightarrow a};
\mathbf{P}^{c\rightarrow a}
\right]
\right),
\end{equation}
where $\{b,c\}=\{xy,xz,yz\}\setminus\{a\}$. Finally, image features are re-injected at each resolution through image-conditioned cross-attention:
\begin{equation}
\mathbf{Z}
\leftarrow
\mathbf{Z}
+
\operatorname{CrossAtt}
\left(
\mathbf{Z},
\mathbf{F}_{\mathrm{img}},
\mathbf{F}_{\mathrm{img}}
\right),
\end{equation}
where $\mathbf{Z}$ stacks the features of the three planes. The refined features are reshaped into three orthogonal planes to obtain $\mathbf{T}_{2s}$.

Across the refinement stages, the Triplane is progressively refined from $64^2$ to $128^2$ and $256^2$, with channel dimensions of $1024$, $768$, and $384$, respectively.

\subsection{Reference Initialization}
\label{app:ref_initialization}
We use fixed spatial references for both Anchor center prediction and Anchor-to-Point prediction. For Anchor centers, we randomly sample $M$ reference centers $\{\mathbf{p}^{0}_{i}\}_{i=1}^{M}$ from the normalized scene space. The reference centers are sampled once at initialization and then kept fixed throughout training.

For Anchor-to-Point prediction, we randomly sample $K$ normalized reference offsets $\{\mathbf{\delta}_{k}\}_{k=1}^{K}$ within the Anchor region. The same set of offsets is sampled once at initialization and shared across all Anchors and training samples. The corresponding reference points are defined as
\begin{equation}
\mathbf{q}_{ik}
=
\mathbf{p}_{i}
+
r_i\mathbf{\delta}_{k}.
\label{eq:point_reference}
\end{equation}

\subsection{Dataset-Specific View Sampling}
\label{app:view_sampling}

The five training datasets have different view organizations, and we therefore use dataset-specific sampling strategies to construct four input views. For ScanNet++ and ARKitScenes, which consist of continuous indoor camera trajectories, we first randomly select a reference view and then sample three additional views within a predefined pose-distance range, avoiding views that are either too close or too far from the reference. For BlendedMVS, we first define minimum and maximum temporal gaps, then randomly select a value within this range to determine the interval between the first and last frames; views are then randomly sampled from within this interval. For CO3D and WildRGB-D, where each sequence contains views of a single object from different viewpoints, we uniformly sample four distinct views from the same sequence.

\subsection{Normalized Geometric Evaluation}
\label{app:geometry_eval}

We evaluate geometry in a common normalized coordinate system defined by the context cameras and ground-truth geometry. The context camera poses determine the orientation and origin of this coordinate system, while the ground-truth point cloud determines its scale. We then transform the ground-truth point cloud and all context cameras into this frame, and normalize the scene to fit within $[-1,1]^3$. For each method, the predicted points are transformed into the same frame using the known correspondence between its context cameras and the ground-truth context cameras. Thus, both ground-truth and predicted points are evaluated in a shared coordinate system.

\subsection{Target-Context Distance}
\label{app:target_context_distance}

We use the \emph{target-context distance} to characterize the viewpoint change in Sec.~\ref{sec:geometry_nvs}. For each target view, it is defined as the Euclidean distance between the target camera center and the nearest context camera center.

As the distance increases, the target view may contain regions that are unseen in all context views. These regions are not observed in the input views. Thus, we compute PSNR only on target pixels that are visible in at least one context view, so that the metric directly reflects the rendering quality of reconstructed regions.

\subsection{Detailed Model Architecture}
\label{app:model_architecture}

We use two model settings in our experiments. The smaller model is used for all ablation studies to reduce computational cost, while the main-table model is used for comparison with prior methods. Their detailed architectures are summarized in Table~\ref{tab:model_architecture}. 

\begin{table}[H]
\centering
\caption{\textbf{Architecture comparison between the smaller model used for ablations and the model used in the main tables.}}
\label{tab:model_architecture}
\small
\setlength{\tabcolsep}{5pt}
\begin{tabular}{lcc}
\toprule
Component & Ablation model & Main-table model \\
\midrule
Input resolution & $256\!\times\!256$ & $256\!\times\!256$ \\
Patch size & $8\!\times\!8$ & $8\!\times\!8$ \\
Image encoder blocks & 6 & 8 \\
Shared Attention blocks & 8 & 12 \\
Triplane resolutions & $64\!\rightarrow\!128\!\rightarrow\!256$ & $64\!\rightarrow\!128\!\rightarrow\!256$ \\
Anchors $\times$ points per Anchor & $4096\!\times\!16$ & $4096\!\times\!16$ \\
Global batch size & 64 & 128 \\
\bottomrule
\end{tabular}
\end{table}

\section{Limitations}
\label{app:limitations}
GeoGSplat relies on geometry supervision during training. Although replacing GT depth with predicted depth maintains stable NVS performance, depth estimation errors may affect geometric accuracy. Moreover, GeoGSplat emphasizes representation efficiency rather than the lowest inference cost, and its computational advantages depend on the number of input views. Finally, the fixed Gaussian budget may limit adaptability to scenes of varying complexity, while the effects of different architectural and training configurations require further investigation.
\end{document}